\documentclass[a4paper]{article}
\usepackage[margin=1.3in]{geometry}

\usepackage{times}

\usepackage{soul}
\usepackage{url}
\usepackage[hidelinks]{hyperref}
\usepackage[utf8]{inputenc}
\usepackage[small]{caption}
\usepackage{graphicx}
\usepackage{amsmath}
\usepackage{booktabs}
\usepackage{amsfonts}       
\usepackage{nicefrac}       
\usepackage{amssymb,amsthm}
\usepackage{natbib}

\newcommand{\R}{\mathbb{R}}
\newcommand{\E}{\mathbb{E}}

\newcommand{\Hcal}{\mathcal{H}}
\newcommand{\inSpace}{\mathcal{X}}
\newcommand{\yb}{\mathbf{y}}
\newcommand{\fb}{\mathbf{f}}
\newcommand{\xb}{\mathbf{x}}
\newcommand{\Xb}{\mathbf{X}}
\newcommand{\ub}{\mathbf{u}}

\newcommand\abs[1][\cdot]{\left\lvert#1\right\rvert}
\newcommand\rank[1]{{\scriptsize(#1)}}

\theoremstyle{remark}
\newtheorem{remark}{Remark}

\begin{document}
	
	\title{Orthogonally Decoupled Variational Fourier Features}
	
	\author{Dario Azzimonti\footnotemark[1] , Manuel Sch\"urch\footnotemark[1] \footnotemark[2] ,  Alessio Benavoli\footnotemark[3] \ and Marco Zaffalon\footnotemark[1]}
	\renewcommand{\thefootnote}{\fnsymbol{footnote}}
	\footnotetext[1]{Dalle Molle Institute for Artificial Intelligence (IDSIA), Lugano, Switzerland}
	\footnotetext[2]{Universit\`a della Svizzera Italiana (USI), Lugano Switzerland}
	\footnotetext[3]{University of Limerick (UL), Limerick, Ireland} 
	\date{}
	
	\maketitle

\begin{abstract}
  Sparse inducing points have long been a standard method to fit Gaussian processes to big data. In the last few years, spectral methods that exploit approximations of the covariance kernel have shown to be competitive. In this work we exploit a recently introduced orthogonally decoupled variational basis to combine spectral methods and sparse inducing points methods. We show that the method is competitive with the state-of-the-art on synthetic and on real-world data. 
\end{abstract}

%

\section{Introduction}


Gaussian processes (GPs) are flexible, non-parametric models often used in regression and classification tasks \citep{rasmussen2006gaussian}. They are probabilistic models and provide both a prediction and an uncertainty quantification. For this reason, GPs are a common choice in different applications see, e.g. \citet{Shahriari.etal2016}, \citet{Santner2018} and \citet{Hennig.eal2015}.  The flexibility of GPs, however, comes with an important computational drawback: training requires the inversion of a $N \times N$ matrix, where $N$ is the size of the training data, resulting in a $O(N^3)$ computational complexity. Many approximations that mitigate this issue have been proposed, see,~\citet{Liu.etal2018} for a review.


Sparse inducing points approaches \citep{quinonero2005unifying} have long been employed for large data \citep{csato2002sparse,seeger2003fast,snelson2006sparse}. The core idea of such methods is to approximate the unknown function with its values at few, $M \ll N$, well-selected input locations called inducing points which leads to a reduced computational complexity of $O(NM^2)$. In \citet{titsias2009variational}  a variational method was introduced that keeps the original GP prior and  approximates the posterior with variational inference. This method guarantees that by increasing the number of inducing inputs the approximate posterior distribution is closer to the full GP posterior in a Kullback-Leibler divergence sense. 

An alternative method for variational inference on sparse GPs was introduced in \citet{ChengBoots2016}, where the authors proposed a variational inference method based on a property of the reproducing kernel Hilbert space (RKHS) associated with the GP. The main idea is to write the variational problem in the RKHS associated with the GP and to parametrize the variational mean and covariance accordingly. This method has been improved in several works \citep{ChengBoots2017,Salimbeni.etal2018} that provide more powerful formulations for the mean and covariance of the variational distribution. In particular, \citet{Salimbeni.etal2018} proposed a powerful orthogonally decoupled basis that allowed for an efficient natural gradient update rule. 

A parallel line of research for sparse GPs studies inter-domain approximations. Such approaches exploit spectral decompositions of the GP kernel and provides low rank approximations of the GP, see \citet{RahimiRecht2007,LazaroGredilla.etal2009,SolinSarkka2014,Hensman.etal2018}. Inter-domain methods replace inducing points variables with more informative inducing features which, usually, do not need to be optimized at training time, thus potentially reducing the computational cost of the approximation method. \citet{Hensman.etal2018} combined the power of an inter-domain approach with the variational setup of \citet{titsias2009variational} in their variational Fourier feature method. 

In this work we combine the flexibility of the orthogonally decoupled RKHS bases introduced in \citet{Salimbeni.etal2018} and the explanatory power of inter-domain approaches to propose a new method for training sparse Gaussian processes. We build a variational distribution parametrized in the mean by an inducing point basis and in the covariance by a variational Fourier features basis. Since variational inference for the basis parameterizing the mean does not require a matrix inversion we can use a large number of inducing points and obtain an approximation close to the true posterior mean. On the other hand, by using a variational Fourier features basis in the covariance, we exploit the higher informative power of such features to obtain better covariance estimates. The orthogonal structure guarantees that the range of the two bases does not overlap. 

Sect.~\ref{sec:IntroGP} reviews Gaussian process for regression and classification and recalls the RKHS property exploited by our approximation. In Sect.~\ref{sec:sparseRKHS} we review the previously proposed techniques for sparse GPs with RKHS bases. We propose our novel technique and we describe the implementation details in Sect.~\ref{sec:ODVFF}. We test our method on synthetic and real data in Sect.~\ref{sec:experiments} and we discuss advantages and drawbacks in Sect.~\ref{sec:discussion}.

\section{Gaussian processes}
\label{sec:IntroGP}

A real valued Gaussian process $f(x) \sim GP(m(x),k(x,x))$, defined on an input space $\inSpace \subset \R^D$, is a stochastic process such that for any $N>0$ the values $\{f(x_i): x_i\in \inSpace \}_{i=1,\ldots, N}$ follow a multivariate normal distribution. It is completely characterized by its mean function $m(x) := \E[f(x)]$ and its covariance kernel $k(x,x^\prime) := \operatorname{Cov}(f(x),f(x^\prime))$. The covariance kernel $k$ is a positive-definite function and a reproducing kernel in an appropriate Hilbert space, see, e.g.,~\citet{BerlinetThomasAgnan2004}. The reproducing property of $k$ implies that there exists a Hilbert space $\Hcal$ such that
\begin{equation*}
k(x,x^\prime) = <\psi(x), \Sigma \psi(x^\prime)>_{\Hcal}, \qquad x,x^\prime \in \inSpace,
\end{equation*}
where $\psi: \inSpace \rightarrow \Hcal$ is a feature map and $\Sigma: \Hcal \rightarrow \Hcal$ is a bounded positive semi-definite self-adjoint operator. Moreover if $m \in \Hcal$, then we can associate a function $\mu \in \Hcal$ such that $m(x) = <\psi(x),\mu>_{\Hcal}$. The couple $\mu, \Sigma$ is a dual representation of a Gaussian process with mean $m$ and covariance $k$ into the space $\mathcal{H}$. Here we follow \citet{Salimbeni.etal2018} and, for simplicity, we denote by $f \sim GP_\Hcal(\mu,\Sigma)$ the dual representation of the $GP(m,k)$ in the Hilbert space $\Hcal$. This notation is only used here to denote that the objects $\mu,\Sigma$ have the dual role of $m, k$, however it does not mean that the GP samples belong to $\Hcal$. 

Let us denote by $\yb = (y_1, \ldots, y_N)$, a vector of $N$ output values and by $\Xb = (x_1, \ldots, x_N )^T$ a matrix of inputs. We consider a likelihood model, not necessarily Gaussian, that factorizes over the outputs, i.e. $p(\yb \mid \fb) = \prod_{i=1}^N p(y_i \mid f(x_i))$, where $\fb = (f(x_i))_{i=1}^N$ and $f\sim GP(m(x),k(x,x^\prime) )$ for a prior mean function $m$ and covariance kernel $k$. The covariance kernel $k$ is often chosen from a parametric family such as the squared exponential or the Mat\'ern family, see \citet{rasmussen2006gaussian}, chapter~4. 
The GP provides a prior distribution for the latent values, i.e. $\fb \sim N(0,K_N)$ and we can use Bayes rule to compute the posterior distribution $p(\fb \mid \yb) = \frac{p(\yb \mid \fb) p(\fb)}{p(\yb)}$. 

Consider now a regression example where we have a training set $\mathcal{D} = \left\{ y_i, x_i \right\}_{i=1}^N=\left(\yb,\Xb\right)$ of $N$ pairs of inputs $x_i\in \R^D$ and noisy scalar outputs $y_i$ generated by adding independent Gaussian noise to a latent function $f(x)$, that is $y_i = f(x_i)+\varepsilon_i$, where $\varepsilon_i\sim \mathcal{N}(0,\sigma_n^2)$. The posterior distribution of the Gaussian process $p(\fb \mid  \yb)$ is Gaussian with mean and covariance given by 
\begin{align*}
m_N(x) &= m(x) + k(x,\Xb) [k(\Xb)+\sigma_{n}^2 I]^{-1}(\yb - m(\Xb)) \\
k_N(x,x^\prime) &= k(x,x^\prime) - k(x,\Xb) [k(\Xb)+\sigma_{n}^2 I]^{-1} k(\Xb,x^\prime)
\end{align*}
where $k(\Xb) = [k(x_i,x_j)]_{i,j =1, \ldots, N} \in \R^{N \times N}$, $k(x,\Xb) = [k(x,x_i)]_{i=1, \ldots, N} \in \R^{1\times N}$, $k(\Xb,x)=k(x,\Xb)^T$ and $m(\Xb) = [m(x_i)]_{i=1,\ldots,N}$.

As mentioned above there exists a Hilbert space $\Hcal$ with inner product $<\cdot, \cdot>_\Hcal$ and a feature map $\psi(x)$ such that $k(x,x^\prime) = <\psi(x), \Sigma\psi(x^\prime)>_\Hcal$. In this case, the operator is $\Sigma = I$ and we further assume that $m(x) = <\psi(x),\mu>_\Hcal$ for some $\mu \in \Hcal$. As an example, we can choose the canonical feature map $\psi(x) = k(x,\cdot)$ and $\Hcal = \Hcal_k$, the RKHS associated with $k$. The posterior mean and covariance above can be rewritten as $m_N(x) = <\psi(x),\mu_N>_{\Hcal}$ and $k_N(x,x^\prime) = <\psi(x),\Sigma_N\psi(x^\prime)>_{\Hcal}$ where 
\begin{align}
\label{eq:muSigma}
\mu_N &:= \mu + \Psi_X [k(\Xb)+\sigma_{n}^2 I]^{-1} (\yb - m(\Xb)) \\
\Sigma_N &:= I-\Psi_X [k(\Xb)+\sigma_{n}^2 I]^{-1} \Psi_X^T.
\end{align}
where $\Psi_X = [\psi(x_1), \ldots, \psi(x_N)]$ and we use the notation
\begin{align*}
\Psi_X^T \psi(x) :&= [<\psi(x_1),\psi(x)>_\Hcal, \ldots, <\psi(x_N),\psi(x)>_\Hcal]^T.
\end{align*}

Note that, in this example, we have $\Psi_X^T \psi(x)= k(\mathbf{X},x)$. 
In the dual representation then the prior GP corresponds to $f \sim GP_\Hcal(\mu,I)$ and the posterior GP to $f \sim GP_\Hcal(\mu_N,\Sigma_N)$ with $\mu_N,\Sigma_N$ as in eq.~\eqref{eq:muSigma}.

From the equations above we can  see that GP training involves the inversion of a matrix of size $N\times N$, thus requiring $O(N^3)$ time. In what follows, we always assume $m(x) \equiv 0$.

\section{Sparse GP and RKHS basis}
\label{sec:sparseRKHS}


In order to reduce the computational cost, we follow here the variational approach introduced in \citet{titsias2009variational} and then generalized to stochastic processes in \citet{Matthews.etal16}. The idea is to approximate the posterior distribution $p(f \mid \yb)$ with a variational distribution $q(f)$. Since the posterior is a GP, the variational distribution should also be a GP. The optimal distribution is then selected as 
\begin{align*}
q &= \arg \min_q KL( q(f) \Vert p(f\mid \yb)) \\
&= \arg\min_q \E_q[\log q(f) - \log p(f \mid \yb)].
\end{align*} 
In \citet{titsias2009variational,Matthews.etal16}, the authors follow the sparse inducing points approach \citep{quinonero2005unifying} and consider $q(f,\ub) = q(f \mid \ub) q(\ub)$ parametrized by $\ub = [f(r_i)]_{i=1, \ldots,M}$ a vector of $M \ll N$ inducing outputs evaluated at inputs $r_1, \ldots, r_M \in \inSpace$. The resulting conditional distribution is 
\begin{align*}
q(f(x) \mid \ub) = GP\bigg(k(x,\ub) K_{\ub,\ub}^{-1} \ub,\quad k(x,x) - k(x,\ub) K_{\ub,\ub}^{-1}k(\ub,x) \bigg).
\end{align*}

The joint distribution $q(f,\ub)$ is optimized, by selecting the variational parameters $b \in \R^M$ and $S \in \R^{M \times M}$ of the variational distribution $q(\ub) = N(b,S)$. In the regression case \citep{titsias2009variational}, the optimal $q(\ub)$ has analytical expressions for its mean and covariance. Stochastic optimization techniques and mini-batch training were developed for regression \citep{hensman2013gaussian,Schuerch.etal2019} and classification \citep{Hensman.eal15}. 

\subsection{Variational problem in the RKHS space}
In the dual view presented in Sect.~\ref{sec:IntroGP} the posterior distribution can be represented as $f\mid \yb \sim GP_\Hcal(\mu_N,\Sigma_N)$. The variational problem can also be represented in this dual form. In particular here the distribution $q$ is represented as $q_{\mathcal{H}}(f) = GP_\Hcal(\mu, \Sigma)$ and we would like to find the distribution $q_{\mathcal{H}}(f)$ that minimizes
\begin{align}
\mathcal{L}(q_{\mathcal{H}}) = - \sum_{i=1}^{N} \E_{q_{\mathcal{H}}(f(x_i))}[ \log p(y_i \mid f(x_i)) ] + KL(q_{\mathcal{H}}(f) \Vert p(f))
\label{eq:lowBound}
\end{align}
where $p(f) \sim GP_{\Hcal}(0,I)$. It can be shown \citep{ChengBoots2016,ChengBoots2017,Salimbeni.etal2018} that $\mathcal{L}(q_{\mathcal{H}}) = KL(q_{\mathcal{H}}(f) \Vert p_{\mathcal{H}}(f \mid \yb) )$ up to a constant. In this dual formulation, the objects $\mu,\Sigma$ are a function and an operator over an Hilbert space respectively and they cannot be optimized directly. In order to optimize $\mu$ and $\Sigma$ we need to choose an appropriate parametrization. \citet{ChengBoots2016}, first proposed the following decoupled decomposition, inspired by eq.~\eqref{eq:muSigma},
\begin{equation}
\mu = \Psi_\alpha a, \qquad \Sigma = I - \Psi_\beta A \Psi_\beta^T,
\label{eq:fullydecNoInv}
\end{equation}
where $\alpha, \beta$ are sets of inducing variables, $a \in \R^{\lvert \alpha \rvert}$, $A \in \R^{\lvert \beta \rvert \times \lvert \beta \rvert}$ are variational parameters and $\Psi_\alpha$ is a basis functions vector defined as $\Psi_\ub= [k(r_1, \cdot), \ldots, k(r_M,\cdot)]^T$, where $u_i =f(r_i) = \alpha_i$ for $\Psi_\alpha$ or $u_i = \beta_i$ for $\Psi_\beta$. If we choose $\alpha = \beta=\ub$, $a = K_{\ub,\ub} b$ and $A= -K_{\ub,\ub}^{-1} (S - K_{\ub,\ub} )K_{\ub,\ub} ^{-1}$ we obtain the standard  \citep{titsias2009variational} result where $b,S$ denote the usual variational parameters to be optimized. 
 

\subsection{Orthogonally decoupled bases}
\label{subsec:ODB}

The decoupled parametrization~\eqref{eq:fullydecNoInv} was shown to be insufficiently constrained in \citet{ChengBoots2017}. If the size of $\beta$ is increased, the basis in \eqref{eq:fullydecNoInv} is not necessarily more expressive and, in particular, the mean does not necessarily improve. 

In~\citet{ChengBoots2017}, the authors further generalized~\eqref{eq:fullydecNoInv} with a hybrid basis that addressed this issue, however the optimization procedure for this basis was shown to be ill-conditioned. Finally, \citet{Salimbeni.etal2018} proposed the following orthogonal bases decomposition.
\begin{align}
\label{eq:orthDecMean}
\mu &= (I - \Psi_\beta K_\beta^{-1}\Psi_\beta^T)\Psi_\gamma a_\gamma +\Psi_\beta a_\beta \\
\Sigma &= I - \Psi_\beta K_\beta^{-1} \Psi_\beta^T + \Psi_\beta K_\beta^{-1} S K_\beta^{-1}\Psi_\beta^T .
\label{eq:orthDecCov}
\end{align}
Pre-multiplying $\Psi_\gamma$ by $(I - \Psi_\beta K_\beta^{-1}\Psi_\beta)$ makes the two bases orthogonal and the optimization problem well-conditioned. In practice the mean function $\mu$ is now decomposed into bases which are no longer overlapping therefore the variational space can be explored more efficiently by the optimizer. 

We propose here to replace the basis functions parametrized by the inducing points $\beta$ with the ones build on RKHS inducing features \citep{Hensman.etal2018}. 

\subsection{Inter-domain approaches}
\label{subsec:interDomain}


An inducing output $u_i$ can be seen as the result of the evaluation functional $L_{r_i}[f]= f(r_i)$, where $u_i = f(r_i)$ and $r_i$ is the inducing input corresponding to the inducing output $u_i$. As long as the resulting random variable $L_{r_i}[f]$ is well defined, we can extend this approach to more general linear functionals; inter-domain sparse GPs are built on this general notion. For example, approaches based on Fourier features \citep{RahimiRecht2007,LazaroGredilla.etal2009} choose the functional $L_\omega[f] = \int f(x) e^{-i\omega x} dx$. 

Here we start by considering only one-dimensional inputs and, by following \citet{Hensman.etal2018}, we restrict the input domain to an interval $[\kappa_0,\kappa_1]\subset \R$. 
%
%
While this might seem like a strong restriction, in practice data is always observed in a finite window and we can select a larger interval $[\kappa_0,\kappa_1]$ that includes all training inputs and the extrapolation region of interest. Moreover, instead of considering an $L^2(\kappa_0,\kappa_1)$ inner product with Fourier features, we consider the RKHS inner product $<\cdot, \cdot>_{\Hcal_k}$, i.e. we consider the RKHS features \citep{Hensman.etal2018}, defined as 

\begin{equation}
\zeta_i := <f, \phi_i>_{\Hcal_k}
\label{eq:VFfeature}
\end{equation}
where $<\cdot, \cdot>_{\mathcal{H}_k}$ is the inner product in $\mathcal{H}_k$, the RKHS induced by the GP covariance kernel $k$, and 
\begin{align}
\nonumber
\Phi = [&\phi_i]_{i=1}^{2M+1} \\
= [&1, \cos(\omega_1(x-\kappa_0)), \ldots, \cos(\omega_M(x-\kappa_0)), \\
&\sin(\omega_1(x-\kappa_0)), \ldots \sin(\omega_M(x-\kappa_0))].
\label{eq:trunFourier}
\end{align}

The covariance between inducing variables and function values and the cross-covariances between inducing variables can be written \citep{Hensman.etal2018} as
\begin{align}
\label{eq:rkhsFeatProperty1}
\operatorname{Cov}(\zeta_i,f(x)) &= \phi_i(x) , \\
\operatorname{Cov}(\zeta_i,\zeta_{i^\prime}) &= <\phi_i,\phi_{i^\prime}>_{\mathcal{H}_k}.
\label{eq:rkhsFeatProperty2}
\end{align}
These properties allow for a fast computation of the kernel matrices as long as the matrix $K_{\phi,\phi}= [<\phi_i,\phi_{i^\prime}>_{\mathcal{H}_k}]_{i,i^\prime=1,\ldots,2M+1}$ is finite and can be computed. Analytical formulae are available \citep{Hensman.etal2018} to compute the inner product above if $k$ is a Mat\'ern kernel with smoothness parameter $\nu=1/2,3/2, 5/2$ defined on an interval $[a, b] \subset \R$.
The method requires the space $\mathcal{F} = span(\Phi)$ to be a subspace of the RKHS $\mathcal{H}_k$. This property is not always true, notable counterexamples \citep{Hensman.etal2018} are the Mat\'ern RKHS on $\R$ and the RBF and Brownian motion kernel on $[0,1]$.  

In the cases where analytical formulae are available, the Gram matrix $K_{\phi,\phi}$ also has the striking property that it can be written as a diagonal matrix plus several rank-one matrices. This allows for more efficient computation of matrix product such as $K_{\phi,\phi}^{-1} K_{\phi,f}$, see \citet{Hensman.etal2018}.

\begin{figure}[ht]
	\includegraphics[width=\linewidth]{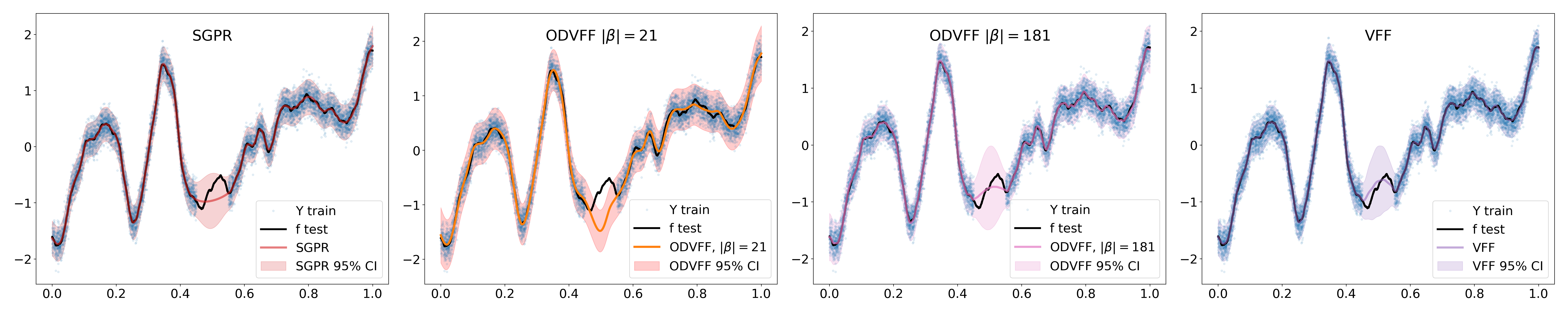}
	\caption{Comparison of different methods, no training data in $[0.45,0.55]$, $\abs[\beta]+\abs[\gamma]=200$.}
	\label{fig:1dExampleComparison}
\end{figure}

\section{Orthogonally decoupled variational Fourier features}
\label{sec:ODVFF}

The orthogonal decomposition in eq.~\eqref{eq:orthDecMean},~\eqref{eq:orthDecCov} only requires an appropriately defined kernel matrix $K_\beta$ and the operators $\Psi_\gamma, \Psi_\beta$; however they do not have to be generated from inducing points. In practice, \citet{Salimbeni.etal2018} shows that the restriction to disjoint sets of inducing points $\gamma$, $\beta$ helps in the optimization procedure, however, the orthogonally decoupled formulation also enforces that the bases related to $\gamma$ and $\beta$ are orthogonal. We can thus exploit this property to obtain a combination of inter-domain basis and inducing points basis which are mutually orthogonal. 

The Orthogonally Decoupled Variational Fourier Features (ODVFF) method considers two sets of variational parameters: a set of inducing points $\gamma = [\gamma_i]_{i=1, \ldots, \abs[\gamma]}$,  and a set of RKHS features $\beta = [ \beta_j]_{j=1, \ldots, \abs[\beta]}$ defined as $\beta_j := <f, \phi_j>_{\mathcal{H}_k}$, with $\phi_j$ defined as in \eqref{eq:trunFourier} for $j=1, \ldots, \abs[\beta]$. We can then build a variational distribution $q$ in the dual space parametrized with $\mu$ and $\Sigma$ defined as

\begin{align} 
\label{eq:orthDecVFFmean}
\mu &= (I - \Psi_\beta K_\beta^{-1}\Psi_\beta^T)\Psi_\gamma a_\gamma +\Psi_\beta a_\beta \\ 
\Sigma &= I - \Psi_\beta K_\beta^{-1} \Psi_\beta^T + \Psi_\beta K_\beta^{-1} S K_\beta^{-1}\Psi_\beta^T .
\label{eq:orthDecVFFvar}
\end{align}
with
\begin{align*}
\Psi_\beta &= [\operatorname{Cov}(\beta_1,f(\cdot)), \ldots, \operatorname{Cov}(\beta_{\lvert \beta \rvert},f(\cdot))]^T  \\
&= [ \phi_1(\cdot), \ldots, \phi_{\lvert \beta \rvert}(\cdot)]^T = \Phi,
\end{align*}
%
and $K_\beta = K_{\phi,\phi} = [\operatorname{Cov}(\beta_i,\beta_{i^\prime})]_{i,i^\prime=1,\ldots, \abs[\beta]}$ is the Gram matrix obtained from the cross-covariance between inducing features. Note that both $\Psi_\beta$ and $K_\beta$ are easily computable by exploiting~\eqref{eq:rkhsFeatProperty1} and~~\eqref{eq:rkhsFeatProperty2}.  $\Psi_\gamma$ is instead obtained from the inducing points basis $[\gamma_i]$ as $\Psi_\gamma(\cdot) = [k(\gamma_1,\cdot), \ldots, k(\gamma_{\abs[\gamma]},\cdot) ]$. 

The distribution above depends on the variational parameters $a_\gamma, a_\beta$ and $S$ which can be either optimized analytically or with stochastic gradient descent and natural gradients, see~\citet{Salimbeni.etal2018b}.  Moreover~\eqref{eq:orthDecVFFmean} and \eqref{eq:orthDecVFFvar} require choosing the hyper-parameters that build the basis $\gamma$ and $\beta$. Note that the parametrization defines $\Sigma$ with hyper-parameters that do not need optimization, in fact the inducing Fourier features $\beta$ are fixed and chosen in advance. This reduces the size of the optimization problem compared to an orthogonally decoupled model with two sets of inducing points. 

We consider here $\abs[\beta] = 2F+1$, $\Phi$ as in \eqref{eq:trunFourier} and the frequencies $\omega_1, \ldots, \omega_F$, are chosen as harmonic on the interval $[\kappa_0,\kappa_1]$, i.e.  $\omega_i  = \frac{2\pi i}{\kappa_1-\kappa_0}$ $i=1, \ldots, F$. Compared to inducing points orthogonally decoupled basis with the same $\abs[\gamma]$ and $\abs[\beta]$, then we only need to optimize $\abs[\gamma]$ inducing parameters. Moreover for the same number of features $F$, inducing RKHS features have been empirically shown to give better fits than inducing points, see \citet{Hensman.etal2018}. Since we parametrize the covariance of the variational distribution with RKHS features, our ODVFF generally obtains better coverage than equivalent orthogonally decoupled inducing points.

\subsection{Multi-dimensional input spaces}

The variational Fourier features used in the previous section were limited to one dimensional input spaces. Here we exploit the extensions introduced in \citet{Hensman.etal2018} to generalize the method to input spaces of any dimension. In particular we look at additive and separable kernels in an hyper-rectangle input space $\inSpace = \prod_{d=1}^{D}[a_d, b_d]$, $a_d,b_d \in \R$.

\subsubsection{Additive kernels}

The first extension to multiple dimensions can be achieved by assuming that the Gaussian process can be decomposed in an additive combination of functions defined on each input dimension. We can write $f(\xb) = \sum_{d=1}^D f_d(x_d)$ where $\xb \in \inSpace$, and $f_d \sim GP(0, k_d(x_d,x_d^\prime))$. This results in an overall process defined as
\begin{equation*}
f \sim GP\left(0, \sum_{d=1}^{D} k_d(x_d,x_d^\prime)\right).
\end{equation*}
The function is decomposed on simple one dimensional functions, therefore we can select a Mat\'ern kernel for each dimension and define $D\abs[\beta]$ features 
\begin{equation*}
\zeta_{i,d} = <\phi_i, f_d>_{\mathcal{H}_d} \quad i=1, \ldots, \abs[\beta], \quad d=1, \ldots, D,
\end{equation*}
where $\mathcal{H}_d$ is the RKHS associated with the kernel of the $d$th dimension. The Gram matrix $K_{\phi,\phi}$ is then the $D\abs[\beta] \times D\abs[\beta]$ block-diagonal matrix where each block of dimension $\abs[\beta] \times \abs[\beta]$ is the one-dimensional Gram matrix.

\subsubsection{Separable kernels}

An alternative approach involves separable kernels, where the process is defined as
\begin{equation*}
f \sim GP\left(0, \prod_{d=1}^{D} k_d(x_d,x_d^\prime)\right).
\end{equation*}

%
In this case we parametrize the basis as the Kronecker product of features over the dimensions, i.e.
\begin{equation*}
\Phi = \bigotimes_{d=1}^D [\phi_{d,1}(x_d), \ldots, \phi_{d,2F+1}(x_d)]^T
\end{equation*}
where  $\phi_{d,1}$ is the first feature for the $d$th dimension. This structure implies that each feature is equal to $\prod_{d=1}^{D} \phi_{d,i}$ and that the inducing variables can be written as $\zeta_i = <~\prod_{d=1}^{D} \phi_{d,i}, f~>_\mathcal{H}$, $i=1, \ldots, \abs[\beta]$. 

The separable structure results in features that are independent across dimensions, i.e. $\operatorname{Cov}(\phi_{d,i},\phi_{d^\prime, j}) =0$  for all $d \neq d^\prime$ and $i,j =1, \ldots, \abs[\beta]$. Moreover, the Gram matrix corresponding to each dimension is the same as the one dimensional case. The overall Gram matrix $K_{\phi,\phi}$ is then a block-diagonal matrix  of size $\abs[\beta]^D \times \abs[\beta]^D$ where each diagonal block of size $\abs[\beta] \times \abs[\beta]$ is the Gram matrix corresponding to a one dimensional problem. As in the one dimensional case the covariance between function values and inducing variables is $\operatorname{Cov}(\zeta_i, f(x)) = \phi_i(x)$. 

Separable kernels scale exponentially in the input dimension $D$. For this reason they are not practically usable for dimensions higher than $3$. Additive kernels on the other hand do not suffer from this problem, however they have a reduced explanatory power because of the strong assumption of independence across dimensions.


\subsection{Variational parameters training}
In order to train the ODVFF model we need to find the variational parameters $a_\gamma, a_\beta, S$ in \eqref{eq:orthDecMean},~\eqref{eq:orthDecCov} such that the quantity in eq.~\eqref{eq:lowBound}, the negative ELBO, is minimized. If the likelihood is Gaussian, the parameters have an analytical expression.
Such expressions however require matrix inversions of sizes $\abs[\gamma] \times\abs[\gamma]$ and $\abs[\beta] \times \abs[\beta]$ as shown in 
 \citet{Salimbeni.etal2018}. 

On the other hand, stochastic mini-batch training requires only the inversion of $K_{\beta}$ at the additional cost of having to numerically optimize the parameters. Since often 
the additional matrix inversion becomes costly, here we only train our models by learning the variational parameters. 
Moreover, here we also exploit the natural parameter formulation introduced in \citet{Salimbeni.etal2018,Salimbeni.etal2018b} for faster and more stable training.


\begin{figure}
	\hspace{0.1in}
	\includegraphics[width=0.8\linewidth]{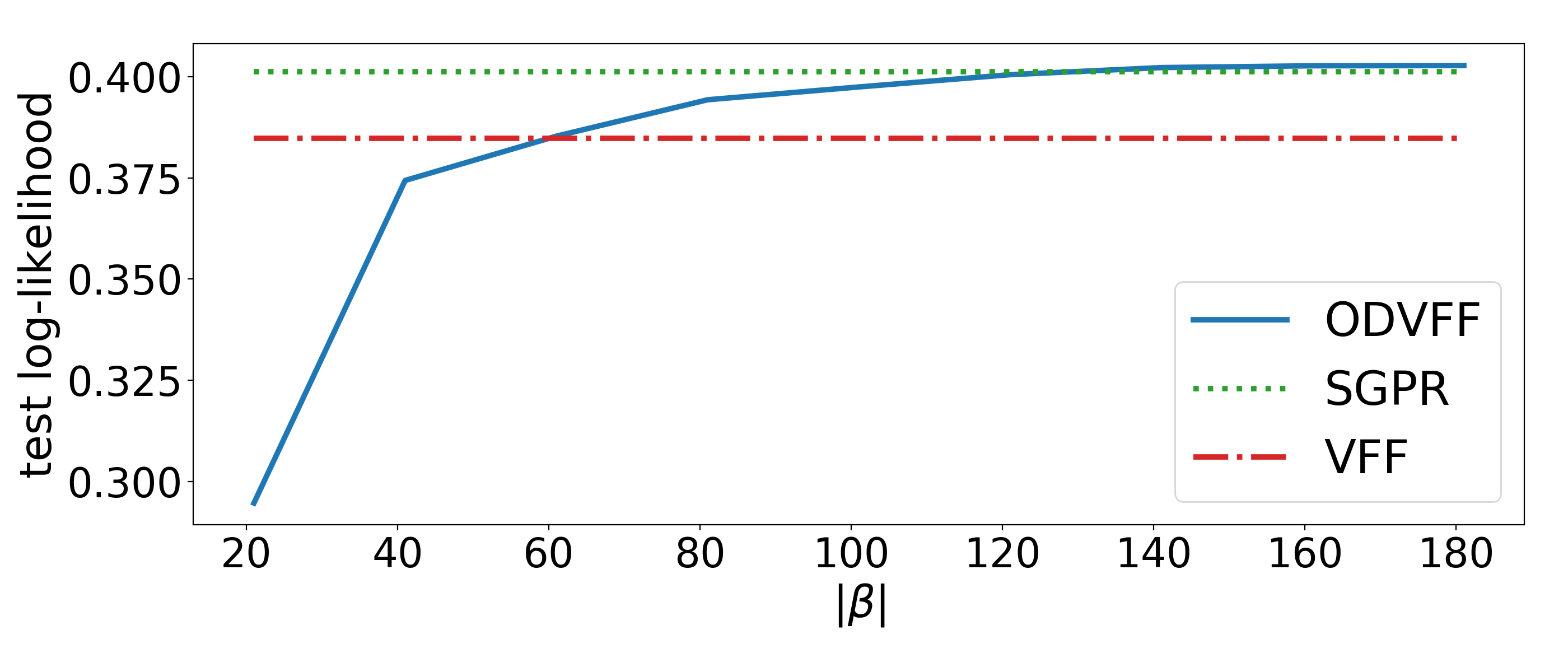}
	\caption{Mean test log-likelihood, $15$ replications. $\abs[\beta]+\abs[\gamma]=200$.}
	\label{fig:1dEx_avgLogLik}
\end{figure}

\subsection{Choice of number of features}
The ODVFF method requires a choice of the number of Fourier features $\abs[\beta]$ and inducing points features $\abs[\gamma]$ to use. This choice involves a trade-off between the cost of inverting the matrix $K_{\beta}$ and that of solving the optimization problem for $\abs[\gamma]$ inducing points.

Figure~\ref{fig:1dExampleComparison} shows a comparison of ODVFF against Variational Fourier Features (VFF) \citep{Hensman.etal2018} and a full-batch variational approach (SGPR) \citep{titsias2009variational}. In this example the data was generated with a GP with mean zero and covariance Mat\'ern with smoothing parameter $\nu=3/2$, length scale $\ell=0.1$ and noise variance $\sigma_n^2 =0.15$. By increasing $\abs[\beta]$ from $21$ (on the left) to $181$ (on the right) the root mean square error (RMSE) of the prediction decreased by $29\%$ and the mean coverage increased from $92\%$ to $96\%$. The confidence intervals at $95\%$ obtained with ODVFF and $\abs[\beta]=181$ become almost indistinguishable from the confidence intervals obtained with SGPR. Note, that SGPR here is trained with the full batch while ODVFF is trained with mini-batches and scales to much larger datasets.

Figure~\ref{fig:1dEx_avgLogLik} shows the average test log-likelihood of ODVFF as a function of $\abs[\beta]$ over $15$ replications of the experiment introduced above. For reference we also plot the average test log-likelihood obtained with VFF and with SGPR.


The improvement in test log-likelihood when $\abs[\beta]$ is increased is mainly driven by a better covariance approximation as indicated by the mean coverage (at $95\%$) on test data which is $98\%$ when $\abs[\beta]=21$ and $96\%$ when $\abs[\beta]=181$.

Table~\ref{tab:synth} shows a comparison between ODVFF and orthogonally decoupled sparse GP (ODVGP) \citep{Salimbeni.etal2018} on a synthetic datasets generated as described in Sect.~\ref{subsec:synthetic}. We train models with $N=50,000$, $\abs[\gamma]+\abs[\beta]=200$ and we consider three choices for $\abs[\beta]$: 11, 99, 189. For each choice we replicate the experiment $10$ times. In this and all following tables bold font highlights the best value. Note that for all dimensions we obtain a better model with a smaller $\abs[\beta]$. Moreover, ODVFF performs better than ODVGP with small $\abs[\beta]$, indicating that the Fourier feature parametrization of the covariance is more powerful. This advantage is reduced when $\beta$ increases as most of the fitting is done with $\gamma$ and the mean parametrization is the same between the two methods. 

{\tiny
	\begin{table}
		\caption{Synthetic data set. Mean test log-likelihood values (rank) over $10$ repetitions, higher values denote a better fit.}
		\label{tab:synth}
		\begin{center}
			\scriptsize
			\begin{tabular}{lrrrr}
				\toprule
				$D$ & & $\abs[\beta]=11$ &        $\abs[\beta]=99$ & $\abs[\beta]=189$  \\
				
				\midrule

				$2$ & \begin{tabular}{@{}c@{}} ODVFF \\ ODVGP\end{tabular}  & \begin{tabular}{@{}c@{}} \textbf{-2.834 \rank{1.4}} \\ -2.875 \rank{4.0}\end{tabular} &  \begin{tabular}{@{}c@{}} -2.896	 \rank{4.9} \\ -2.845 \rank{2.2}\end{tabular} & \begin{tabular}{@{}c@{}}
					-2.901 \rank{5.8} \\
					-2.847 \rank{2.7}
				\end{tabular} \\[0.2cm]
				
				
				$4$ & \begin{tabular}{@{}c@{}} ODVFF \\ ODVGP\end{tabular} & \begin{tabular}{@{}c@{}} -1.763 \rank{2.8} \\-1.780 \rank{3.8} \end{tabular} & \begin{tabular}{@{}c@{}} -1.796 \rank{4.8} \\ \textbf{-1.751 \rank{1.1}} \end{tabular} & \begin{tabular}{@{}c@{}} -1.923 \rank{5.9} \\ -2.02 \rank{2.6} \end{tabular} \\[0.2cm]
				

				
				$8$ & \begin{tabular}{@{}c@{}} ODVFF \\ ODVGP\end{tabular} & \begin{tabular}{@{}c@{}} \textbf{-2.293 \rank{1.0}} \\ -2.356 \rank{2.0}\end{tabular} & \begin{tabular}{@{}c@{}} -3.630 \rank{3.0} \\ -4.887 \rank{4.5}\end{tabular} & \begin{tabular}{@{}c@{}}
					-4.660 \rank{4.3} \\
					-6.173 \rank{5.7}
				\end{tabular} \\
				

				
			\end{tabular}
		\end{center}
	\end{table}
}

\section{Experiments}
\label{sec:experiments}

In this section we compare ODVFF with ODVGP, the stochastic variational inference method (SVGP) in \citet{hensman2013gaussian} and with the full-batch variational approach (SGPR) in \citet{titsias2009variational}. We implement our experiments in GPflow \citep{GPflow2017}. We estimate the kernel hyper-parameters and the inducing points locations with the Adam \citep{KingmaBa2015} implementation in Tensorflow. The variational parameters are estimated with Adam and the natural gradient descent method described in \citet{Salimbeni.etal2018b} for SVGP and ODVGP, ODVFF respectively. 

\subsection{Synthetic data}
\label{subsec:synthetic}

We test our method on synthetic data generated from Gaussian process realizations. We consider $D=1, 5, 10$ and two different training setups with $N=50,100 \times 10^3$. In all cases the training data is generated as realizations of a zero mean GP with ARD Mat\'ern kernel with smoothing parameter $\nu=3/2$, length scale $\ell=0.1$ for each dimension and variance $\sigma_o^2=1$. The observations are noisy with independent Gaussian noise with variance $\sigma_n^2 =0.2$. We consider a prior GP with mean zero and additive Mat\'ern covariance kernel with $\nu=3/2$. For all dimensions and for all $N$ we optimize the inducing points locations and the hyper-parameters. We fix the total number of inducing parameters to $\abs[\gamma] +\abs[\beta] = 100$ and we consider two scenarios: $\abs[\beta]=10$ and $\abs[\beta]=50$. 

Tables~\ref{tab:synth1D}, \ref{tab:synth5D}, \ref{tab:synth10D} show the test log-likelihood values obtained on $1000$ test data points for $D=1,5,10$ respectively. A comparison of the RMSE and the mean coverage values at $95\%$ are reported in appendix. SGPR should be considered as a benchmark as it is the only full-batch method. The other three methods are trained on mini-batches of size $500$ with $8000$ iterations. SVGP and SGPR do not depend on $\abs[\beta]$ but on the overall number of inducing points, thus their likelihood values are simply repeated in those columns. SGPR could not be run in the case $N=100 \times 10^3$ due to memory limitations. ODVFF shows better performance than all other methods with $N=100\times 10^3$ in dimensions $D=5,10$ while it is below SVGP when $D=1$. The difference in RMSE is not very large, while the test mean coverages are significantly different. This further reinforces the idea that a variational Fourier feature parametrization for the covariance allows for better fits.

{\tiny
\begin{table}
	\caption{Synthetic data sets, $D=1$. Average test log-likelihood values (rank) over $20$ repetitions, higher values denote a better fit.}
	\label{tab:synth1D}
	\begin{center}
		\scriptsize
		\begin{tabular}{lrrrr}
			& \multicolumn{2}{c}{$N=50000$} & \multicolumn{2}{c}{$N=100000$} \\
			\cmidrule(lr){2-3}
			\cmidrule(lr){4-5}
			& $\abs[\beta]=10$ &        $\abs[\beta]=50$ &        $\abs[\beta]=10$ &        $\abs[\beta]=50$ \\
			
			\midrule
			ODVFF &   0.085 \rank{2.0} &  \textbf{0.147 \rank{1.5}}&  0.049 \rank{2.20}&  0.040 \rank{2.1} \\
			ODVGP &  0.084 \rank{2.0} &  0.135 \rank{1.7} &  0.090 \rank{2.15}& -0.022 \rank{2.7}\\
			SVGP  &  \textbf{0.090 \rank{2.0}} & 0.090 \rank{2.8} &  \textbf{0.112 \rank{1.65}} & \textbf{0.112 \rank{1.2}} \\
			SGPR & \multicolumn{2}{c}{0.156}  &   \multicolumn{2}{c}{$-$} \\
		\end{tabular}
	\end{center}
\end{table}

}



%

{\tiny
\begin{table}
	\caption{Synthetic data sets, $D=5$. Average test log-likelihood values (rank) over $20$ repetitions, higher values denote a better fit.}
	\label{tab:synth5D}
	\begin{center}
		\scriptsize
		\begin{tabular}{lrrrr}
			& \multicolumn{2}{c}{$N=50000$} & \multicolumn{2}{c}{$N=100000$} \\
			\cmidrule(lr){2-3}
			\cmidrule(lr){4-5}
			& $\abs[\beta]=10$ &        $\abs[\beta]=50$ &        $\abs[\beta]=10$ &        $\abs[\beta]=50$ \\
			\midrule
			ODVFF & \textbf{-2.421 \rank{1.0}} & -4.450 \rank{2.00} & \textbf{-6.278 \rank{1.35}} & \textbf{-4.611 \rank{1.10}} \\
			ODVGP & -2.518 \rank{2.0} & -4.653 \rank{2.25} & -7.885 \rank{2.90} & -5.767 \rank{2.15} \\
			SVGP  & -2.927 \rank{3.0} & \textbf{-2.927 \rank{1.75}} & -6.396 \rank{1.75} & -6.396 \rank{2.75} \\
			SGPR  & \multicolumn{2}{c}{-1.694} &    \multicolumn{2}{c}{$-$} \\
		\end{tabular}
	\end{center}
\end{table}

}


{\tiny
\begin{table}
	\caption{Synthetic data sets, $D=10$. Average test log-likelihood values (rank) over $20$ repetitions, higher values denote a better fit.} 
	\label{tab:synth10D}
	\begin{center}
		\scriptsize
		\begin{tabular}{lrrrr}
			
			& \multicolumn{2}{c}{$N=50000$} & \multicolumn{2}{c}{$N=100000$} \\
			\cmidrule(lr){2-3}
			\cmidrule(lr){4-5}
			& $\abs[\beta]=10$ &        $\abs[\beta]=50$ &        $\abs[\beta]=10$ &        $\abs[\beta]=50$ \\
			
			\midrule
			ODVFF & \textbf{-2.989 \rank{1.0}} & \textbf{-2.866 \rank{1.1}} & \textbf{-2.640 \rank{1.0}} & \textbf{-2.356 \rank{1.0}} \\
			ODVGP & -3.164 \rank{2.0} & -7.253 \rank{2.9} & -3.233 \rank{2.0} & -9.451 \rank{3.0} \\
			SVGP  & -3.998 \rank{3.0} & -3.998 \rank{2.0} & -4.076 \rank{3.0} & -4.076 \rank{2.0} \\
			SGPR  & \multicolumn{2}{c}{-1.842} & \multicolumn{2}{c}{$-$} \\
			
		\end{tabular}
	\end{center}
\end{table}
}



\subsection{Benchmarks}

In this section we benchmark the methods on $7$ regression datasets from the UCI Machine Learning repository\footnote{https://archive.ics.uci.edu/ml/index.php} and $10$ datasets from the Penn Machine Learning Benchmark (PMLB) \citep{Olson2017PMLB}. We consider a prior GP with mean zero and additive kernel from the Mat\'ern family with $\nu=3/2$. For all datasets we fix $\abs[\gamma]=300$, $\abs[\beta]=100$, except for kegg\_directed ($\abs[\gamma]=150$, $\abs[\beta]=50$) and sgemmProd ($\abs[\gamma]=120$, $\abs[\beta]=80$). 

Table~\ref{tab:uciLogLik} shows the test log-likelihood values obtained on test data selected randomly as $10\%$ of the original dataset with UCI data on the top part and PMLB data in the bottom. Each experiment is repeated $5$ times with different train/test splits. The hyper-parameters, including the inducing points locations, are estimated by maximizing the likelihood of the model for all experiments except for the airline dataset where the inducing points locations are chosen with k-means from the training data. For each model we run the stochastic optimizer for $10000$ iterations with mini-batches of size $400$. We use the same learning rate for all hyper-parameters and a different learning rate in the natural gradient descent for the variational parameters. 

{\scriptsize
\begin{table*}
	\caption{Average test log-likelihood (test RMSE) on UCI/PMLB datasets. Higher (lower) values denote a better fit.}
	\label{tab:uciLogLik}
	\begin{center}
		\footnotesize
\begin{tabular}{lrrr}
	\toprule[\heavyrulewidth]
	
	  & ODVFF &  ODVGP &   SVGP \\
	  
	  \midrule
	   
	  {\scriptsize 3droad (D=3, N=434,874)}  &  \textbf{-1.174 (0.78)} & -1.558 (0.79) & -1.224 (0.79) \\
	  {\scriptsize airline (D=8, N=1,052,631)} & \textbf{-1.881 (0.91)} & -2.546 (0.92)  &	-2.789 \textbf{(0.91)} \\  
	  {\scriptsize bike (D=12, N=17,379)}  &  -0.859 \textbf{(0.56)} & \textbf{-0.848 (0.56)} & -0.950 (0.57) \\
	  {\scriptsize kegg\_directed (D=19, N=53,413)} &  0.169 \textbf{(0.09)} & -0.074 (0.11) &  \textbf{0.757} (0.10) \\
	  {\scriptsize protein (D=9, N=45,730)} &  \textbf{-1.221} (0.82) & \textbf{-1.221 (0.81)} & -1.348 \textbf{(0.81)} \\
	  {\scriptsize sgemmProd (D=14, N=241,600)} & -1.180 \textbf{(0.63)} &	\textbf{-1.047} (0.64) &	-7.175 \textbf{(0.63)} \\
	  {\scriptsize tamilelectric (D=2, N=45,781)} &  \textbf{-1.449 (0.99)} & -1.450 \textbf{(0.99)} & -1.492 \textbf{(0.99)} \\
  	  \midrule[\heavyrulewidth] 
  	 
  	  	{\scriptsize 215\_2dplanes (D=10, N=40,768)} &  \textbf{-0.882 (0.54)} &  -0.888 \textbf{(0.54)} &  -2.390 (0.63) \\
  	  	{\scriptsize 537\_houses (D=8, N=20,640)} &  \textbf{-0.685 (0.48)}&  -0.693 \textbf{(0.48)} &  -0.707 \textbf{(0.48)} \\
  	  	{\scriptsize BNG\_breastTumor (D=9, N=116,640)} &  \textbf{-1.479} (0.93) & -11.894 (0.93) & -19.157 \textbf{(0.93)} \\
  	  	{\scriptsize BNG\_echoMonths (D=9, N=17,496)} &  -1.149 (0.75) &  \textbf{-1.133} (0.75) &  -1.233 \textbf{(0.74)} \\
  	  	{\scriptsize BNG\_lowbwt (D=9, N=31,104)} &  -0.997 (0.65) &  \textbf{-0.980 (0.64)} &  -1.062 (0.64) \\
  	  	{\scriptsize BNG\_pbg (D=18, N=1,000,000)} &  \textbf{-1.200 (0.80)} &  -2.276 (0.80) &  -5.092 (0.80) \\
  	  	{\scriptsize BNG\_pharynx (D=10, N=1,000,000)} &  \textbf{-1.104 (0.72)} &  -3.018 (0.73) & -11.410 \textbf{(0.72)} \\
  	  	{\scriptsize BNG\_pwLinear (D=10, N=177,147)} &  \textbf{-1.060} \textbf{(0.69)} &  -2.720 \textbf{(0.69)} & -10.911 \textbf{(0.69)} \\
  	  	{\scriptsize fried (D=10, N=40,768)} &  -0.338 \textbf{(0.34)} &  \textbf{-0.334}\textbf{ (0.34)} &  -0.358 \textbf{(0.34)} \\
  	  	{\scriptsize mv (D=10, N=40,768)} &  \textbf{-0.544} \textbf{(0.42)} &  \textbf{-0.544 (0.42)} &  -0.576 \textbf{(0.42)} \\
  	  	
  	  \midrule[\heavyrulewidth] 
  	  
  	  \textbf{Mean rank logL (RMSE)} &
  	  \textbf{1.448} (2.000) & 1.634 (2.241) & 2.914 \textbf{(1.759)} \\
	  
	  \bottomrule
\end{tabular}
\end{center}
\end{table*}

}


In this benchmark the ODVFF method provides either the best or the second best value in test log-likelihood. Note that there is little variation between splits: the ratio between standard deviation and average log-likelihood over the splits, across all datasets is 0.023 for ODVFF, 0.041 for ODVGP and 0.034 for SVGP. Note that while there are some datasets for which this ratio is higher, the worst ratio is achieved by ODVGP on the dataset BNG\_breastTumor and it is equal to 0.46. 
In particular, in the datasets with large $N$, such as airline, BNG\_pbg and BNG\_pharynx, ODVFF greatly outperforms ODVGP and SVGP in terms of test log-likelihood. The mean test RMSE values, reported in parenthesis, are not very different between the methods, however ODVFF is consistently better than ODVGP.  

\section{Discussion}
\label{sec:discussion}

In this work we introduced a novel algorithm to train variational sparse GP. We consider a parametrization of the variational posterior approximation which is based on two orthogonal bases: an inducing points basis and a Fourier feature basis. This approach allows to exploit the accuracy in RMSE obtained with inducing points methods and allows for a better parametrization of the posterior covariance by exploiting the higher explanatory power of variational Fourier features. Our method also inherits the limitation of variational Fourier features to the kernels for which analytical expressions for the Gram matrix are available. The method compares favorably with respect to full inducing points orthogonally decoupled methods and retains the computational stability of this method.

\small
\bibliographystyle{apalike}
\bibliography{biblio_ODVFF}

\begin{thebibliography}{}

\bibitem[Berlinet and Thomas-Agnan, 2004]{BerlinetThomasAgnan2004}
Berlinet, A. and Thomas-Agnan, C. (2004).
\newblock {\em Reproducing kernel Hilbert spaces in probability and
  statistics}.
\newblock Kluwer Academic Publishers.

\bibitem[Cheng and Boots, 2016]{ChengBoots2016}
Cheng, C.-A. and Boots, B. (2016).
\newblock Incremental variational sparse gaussian process regression.
\newblock In Lee, D.~D., Sugiyama, M., Luxburg, U.~V., Guyon, I., and Garnett,
  R., editors, {\em Advances in Neural Information Processing Systems 29},
  pages 4410--4418. Curran Associates, Inc.

\bibitem[Cheng and Boots, 2017]{ChengBoots2017}
Cheng, C.-A. and Boots, B. (2017).
\newblock Variational inference for gaussian process models with linear
  complexity.
\newblock In Guyon, I., Luxburg, U.~V., Bengio, S., Wallach, H., Fergus, R.,
  Vishwanathan, S., and Garnett, R., editors, {\em Advances in Neural
  Information Processing Systems 30}, pages 5184--5194. Curran Associates, Inc.

\bibitem[Csat{\'o} and Opper, 2002]{csato2002sparse}
Csat{\'o}, L. and Opper, M. (2002).
\newblock Sparse online gaussian processes.
\newblock {\em Neural computation}, 14(3):641--668.

\bibitem[Hennig et~al., 2015]{Hennig.eal2015}
Hennig, P., Osborne, M.~A., and Girolami, M. (2015).
\newblock Probabilistic numerics and uncertainty in computations.
\newblock {\em Proceedings of the Royal Society A: Mathematical, Physical and
  Engineering Sciences}, 471(2179):20150142.

\bibitem[Hensman et~al., 2018]{Hensman.etal2018}
Hensman, J., Durrande, N., and Solin, A. (2018).
\newblock Variational {F}ourier features for {G}aussian processes.
\newblock {\em Journal of Machine Learning Research}, 18(151):1--52.

\bibitem[Hensman et~al., 2013]{hensman2013gaussian}
Hensman, J., Fusi, N., and Lawrence, N.~D. (2013).
\newblock Gaussian processes for big data.
\newblock In {\em Conference for Uncertainty in Artificial Intelligence}.

\bibitem[Hensman et~al., 2015]{Hensman.eal15}
Hensman, J., Matthews, A., and Ghahramani, Z. (2015).
\newblock {Scalable Variational Gaussian Process Classification}.
\newblock In Lebanon, G. and Vishwanathan, S. V.~N., editors, {\em Proceedings
  of the Eighteenth International Conference on Artificial Intelligence and
  Statistics}, volume~38 of {\em Proceedings of Machine Learning Research},
  pages 351--360, San Diego, California, USA. PMLR.

\bibitem[Kingma and Ba, 2015]{KingmaBa2015}
Kingma, D.~P. and Ba, J. (2015).
\newblock Adam: {A} method for stochastic optimization.
\newblock In {\em 3rd International Conference on Learning Representations,
  {ICLR} 2015, San Diego, CA, USA, May 7-9, 2015, Conference Track
  Proceedings}.

\bibitem[L\'{a}zaro-Gredilla and Figueiras-Vidal,
  2009]{LazaroGredilla.etal2009}
L\'{a}zaro-Gredilla, M. and Figueiras-Vidal, A. (2009).
\newblock Inter-domain gaussian processes for sparse inference using inducing
  features.
\newblock In Bengio, Y., Schuurmans, D., Lafferty, J.~D., Williams, C. K.~I.,
  and Culotta, A., editors, {\em Advances in Neural Information Processing
  Systems 22}, pages 1087--1095. Curran Associates, Inc.

\bibitem[Liu et~al., 2018]{Liu.etal2018}
Liu, H., Ong, Y.-S., Shen, X., and Cai, J. (2018).
\newblock {When Gaussian Process Meets Big Data: A Review of Scalable GPs}.
\newblock {\em arXiv:1807.01065}.

\bibitem[Matthews et~al., 2016]{Matthews.etal16}
Matthews, A. G. d.~G., Hensman, J., Turner, R., and Ghahramani, Z. (2016).
\newblock On sparse variational methods and the {Kullback-Leibler} divergence
  between stochastic processes.
\newblock In Gretton, A. and Robert, C.~C., editors, {\em Proceedings of the
  19th International Conference on Artificial Intelligence and Statistics},
  volume~51 of {\em Proceedings of Machine Learning Research}, pages 231--239,
  Cadiz, Spain. PMLR.

\bibitem[Matthews et~al., 2017]{GPflow2017}
Matthews, A. G. d.~G., {van der Wilk}, M., Nickson, T., Fujii, K.,
  {Boukouvalas}, A., {Le{\'o}n-Villagr{\'a}}, P., Ghahramani, Z., and Hensman,
  J. (2017).
\newblock {GP}flow: A {G}aussian process library using {T}ensor{F}low.
\newblock {\em Journal of Machine Learning Research}, 18(40):1--6.

\bibitem[Olson et~al., 2017]{Olson2017PMLB}
Olson, R.~S., La~Cava, W., Orzechowski, P., Urbanowicz, R.~J., and Moore, J.~H.
  (2017).
\newblock Pmlb: a large benchmark suite for machine learning evaluation and
  comparison.
\newblock {\em BioData Mining}, 10(1):36.

\bibitem[Qui{\~n}onero-Candela and Rasmussen, 2005]{quinonero2005unifying}
Qui{\~n}onero-Candela, J. and Rasmussen, C.~E. (2005).
\newblock A unifying view of sparse approximate gaussian process regression.
\newblock {\em Journal of Machine Learning Research}, 6(Dec):1939--1959.

\bibitem[Rahimi and Recht, 2007]{RahimiRecht2007}
Rahimi, A. and Recht, B. (2007).
\newblock Random features for large-scale kernel machines.
\newblock In {\em Proceedings of the 20th International Conference on Neural
  Information Processing Systems}, NIPS'07, pages 1177--1184, USA. Curran
  Associates Inc.

\bibitem[Rasmussen and Williams, 2006]{rasmussen2006gaussian}
Rasmussen, C.~E. and Williams, C.~K. (2006).
\newblock {\em Gaussian processes for machine learning}.
\newblock MIT press Cambridge.

\bibitem[Salimbeni et~al., 2018a]{Salimbeni.etal2018}
Salimbeni, H., Cheng, C.-A., Boots, B., and Deisenroth, M. (2018a).
\newblock Orthogonally decoupled variational gaussian processes.
\newblock In Bengio, S., Wallach, H., Larochelle, H., Grauman, K.,
  Cesa-Bianchi, N., and Garnett, R., editors, {\em Advances in Neural
  Information Processing Systems 31}, pages 8725--8734. Curran Associates, Inc.

\bibitem[Salimbeni et~al., 2018b]{Salimbeni.etal2018b}
Salimbeni, H., Eleftheriadis, S., and Hensman, J. (2018b).
\newblock Natural gradients in practice: Non-conjugate variational inference in
  gaussian process models.
\newblock In Storkey, A. and Perez-Cruz, F., editors, {\em Proceedings of the
  Twenty-First International Conference on Artificial Intelligence and
  Statistics}, volume~84 of {\em Proceedings of Machine Learning Research},
  pages 689--697, Playa Blanca, Lanzarote, Canary Islands. PMLR.

\bibitem[Santner et~al., 2018]{Santner2018}
Santner, T.~J., Williams, B.~J., and Notz, W.~I. (2018).
\newblock {\em The Design and Analysis of Computer Experiments}.
\newblock Springer New York, New York, NY.

\bibitem[Sch\"urch et~al., 2019]{Schuerch.etal2019}
Sch\"urch, M., Azzimonti, D., Benavoli, A., and Zaffalon, M. (2019).
\newblock {Recursive Estimation for Sparse Gaussian Process Regression}.
\newblock {\em arXiv:1905.11711}.

\bibitem[Seeger et~al., 2003]{seeger2003fast}
Seeger, M., Williams, C., and Lawrence, N. (2003).
\newblock Fast forward selection to speed up sparse gaussian process
  regression.
\newblock In {\em Artificial Intelligence and Statistics 9}, number
  EPFL-CONF-161318.

\bibitem[{Shahriari} et~al., 2016]{Shahriari.etal2016}
{Shahriari}, B., {Swersky}, K., {Wang}, Z., {Adams}, R.~P., and {de Freitas},
  N. (2016).
\newblock Taking the human out of the loop: A review of bayesian optimization.
\newblock {\em Proceedings of the IEEE}, 104(1):148--175.

\bibitem[Snelson and Ghahramani, 2006]{snelson2006sparse}
Snelson, E. and Ghahramani, Z. (2006).
\newblock Sparse gaussian processes using pseudo-inputs.
\newblock In {\em Advances in Neural Information Processing Systems}, pages
  1257--1264.

\bibitem[Solin and Särkkä, 2014]{SolinSarkka2014}
Solin, A. and Särkkä, S. (2014).
\newblock Hilbert space methods for reduced-rank {G}aussian process regression.
\newblock {\em arXiv preprint arXiv:1401.5508}.

\bibitem[Titsias, 2009]{titsias2009variational}
Titsias, M. (2009).
\newblock Variational learning of inducing variables in sparse gaussian
  processes.
\newblock In {\em Artificial Intelligence and Statistics}, pages 567--574.

\end{thebibliography}

\appendix

\section{More on the coupled basis}
The idea behind the coupled basis introduced in \citet{ChengBoots2016} is to exploit the representation in eq.~\eqref{eq:muSigma} and to define the variational posterior with a similar structure. 

The starting point is the coupled representation
\begin{align}
&\mu = \Psi_\alpha a_\alpha \ &\Sigma = I - \Psi_\alpha A \Psi_\alpha^T,
\label{eq:coupled}
\end{align}
where $\alpha$ are $M=\abs[\alpha]$ inducing variables, $a_\alpha \in \R^{M}$, $A \in \R^{M \times M}$ and $\Psi_\alpha$ is a basis functions vector which depends on $\alpha$. The most basic choice is  $\Psi_\alpha= [k(u_1, \cdot), \ldots, k(u_M,\cdot)]^T$, where $u_1, \ldots, u_M \in \R^D$ are inducing points. With this choice if we select
\begin{align}
&a_\alpha = K_{uu}^{-1}b \ &A= - K_{uu}^{-1}(S - K_{uu})K_{uu}^{-1}
\label{eq:dtcParams}
\end{align}
where $b$ and $S$ are variational parameters to be optimized, we go back to the Titsias model 
\begin{align*}
&\mu = \Psi_u K_{uu}^{-1}b \ &\Sigma = I + \Psi_u K_{uu}^{-1}(S- K_{uu})K_{uu}^{-1}\Psi_u^T
\end{align*}
corresponding to
\begin{align*}
m(x) &= k(x,U) K_{uu}^{-1}b \\
k(x,y) &= k(x,y) + k(x,U)K_{uu}^{-1}(S- K_{uu})K_{uu}^{-1}k(U,y).
\end{align*}
If we take $S= [K_{uu}^{-1}+ \tfrac{1}{\sigma_n^ 2}K_{uu}^{-1}K_{uf}K_{fu}K_{uu}^{-1}]$ and $b=\tfrac{1}{\sigma_n^ 2} S K_{uf}y$ we get the analytical expressions for the variational posterior of Titsias model.

The analytical expressions above results from the minimization of the KL divergence between $q(f_X,f_U)$ (the variational approximation) and $p(f_X, f_U \mid y)$, the actual posterior distribution. The Titsias approximation is $q(f_X,f_U) = p(f_X \mid f_U)q(f_U)$ where $p(f_X \mid f_U) = N(f_X \mid K_{Xu} K_{uu}^{-1}f_U , K_X - K_{Xu} K_{uu}^{-1} K_{uX} )$.

The parameterization in eq.~\eqref{eq:dtcParams} makes the RKHS model equivalent to the standard DTC model, however the main advantage is that the distribution in the RKHS space $q(f) = N(f \mid \mu, \Sigma)$ is equivalent to $q(f_X, f_U)$. 

\begin{remark}
	The RKHS distribution $q(f) = N(f \mid \mu, \Sigma)$ is equivalent to $q(f_X, f_U)$, in particular
	{\small
		\begin{equation}
		q(f) = p(f_X \mid f_U) q(f_U) \abs[ K_U ]^{1/2} \abs[K_X - K_{Xu} K_{uu}^{-1}K_{uX}]^{1/2}.
		\label{eq:equivCoupled}
		\end{equation}}
	\label{rem:equivCoupled}
\end{remark}
This equivalent formulation allows to write 
\begin{align}
\nonumber
\max\mathcal{L}(q(f)) &= \max \int q(f) \log\left( \frac{p(y \mid f)p(f)}{q(f)} \right) df \\
&= \max \int q(f) \left( \sum_{i=1}^{N} \log p(y_i\mid f) +\log\frac{p(f)}{q(f)} \right) df
\label{eq:lowBoundDecomp}
\end{align}
which can be easily optimized with stochastic optimizers because is a sum over $N$ observations. 

\begin{proof}[Proof of remark~\ref{rem:equivCoupled}]
	Consider $q(f) = N(f \mid \mu, \Sigma)$ with $\mu = \Psi_u K_{uu}^{-1}m$ and $\Sigma = I+\Psi_u K_{uu}^{-1} (S - K_{uu}) K_{uu}^{-1} \Psi_u^T$. We have 
	\begin{align*}
	-\log q(f) = \frac{1}{2} \log \abs[\Sigma] + \frac{1}{2} (f-\mu)^T \Sigma^{-1} (f-\mu),
	\end{align*}
	moreover 
	{\small
		\begin{align*}
		\log \abs[\Sigma] &= \log \abs[I+\Psi_u K_{uu}^{-1} (S - K_{uu}) K_{uu}^{-1} \Psi_u^T]  \\
		&=  \log \abs[(S - K_{uu})^{-1}+K_{uu}^{-1} \Psi_u^T\Psi_u K_{uu}^{-1}]\abs[(S - K_{uu})] \\
		&= \log \abs[ (K_{uu} - K_{uu}(S - K_{uu}+K_{uu})^{-1}K_{uu} )^{-1}]\abs[(S - K_{uu})] \\
		&= \log \abs[ ( K_{uu} - K_{uu}S^{-1}K_{uu} )^{-1} ]\abs[(S - K_{uu})] \\
		&= \log \frac{\abs[S - K_{uu}]}{\abs[ ( K_{uu} - K_{uu}S^{-1}K_{uu} )]} = \log \frac{\abs[S - K_{uu}]}{\abs[K_{uu}S^{-1}]\abs[ S - K_{uu} ]} \\
		&= \log \frac{\abs[S]}{\abs[K_{uu}]}
		\end{align*}
	}
	and 
	
	\begin{align*}
	\Sigma^{-1} &= (I+\Psi_u K_{uu}^{-1} (S - K_{uu}) K_{uu}^{-1} \Psi_u^T)^{-1}  \\
	&= I-\Psi_u K_{uu}^{-1} ((S - K_{uu})^{-1} \\
	&\quad+ K_{uu}^{-1} \Psi_u^T \Psi_u K_{uu}^{-1} )^{-1} K_{uu}^{-1} \Psi_u^T \\
	&= I-\Psi_u K_{uu}^{-1} ((S - K_{uu})^{-1} + K_{uu}^{-1} )^{-1} K_{uu}^{-1} \Psi_u^T \\
	&= I-\Psi_u K_{uu}^{-1} ( K_{uu} - K_{uu}S^{-1}K_{uu} ) K_{uu}^{-1} \Psi_u^T \\
	&= I - \Psi_u ( K_{uu}^{-1} - S^{-1} ) \Psi_u^T
	\end{align*}
	
	Finally we also have that $f = f_\parallel + f_\perp$ where $f_\parallel = \Psi_u K_{uu}^{-1} f_u$ and $f_\perp$ is such that $f_\perp = (I -\Psi_u K_{uu}^{-1}\Psi_u^T)f_\perp = (I-P_u)f_\perp = N_u f_\perp$. We can find $b$ such that $f_\perp = \Psi_X b$ which implies $\Psi_X P_u \Psi_X^T b = 0$ (because $f_\perp$ is in the null space of $P_u$). This means that $b$ is in the null space of $\hat{K_u} = \Psi_X P_u \Psi_X^T$ and $\hat{N}b =b$, therefore we have
	
	\begin{align*}
	f_X - K_{X,u} K_{uu}^{-1}f_u &= \Psi_X^T (I - \Psi_u K_{uu}^{-1} \Psi_u^T) f \\
	&= \Psi_X^T f_\perp = \Psi_X^T N_u f_\perp \\
	&= \Psi_X^T N_u \Psi_X b = \Psi_X^T N_u \Psi_X \hat{N} b \\
	&= (K_X - K_{X,u} P_u K_{u,X}) \hat{N} b.
	\end{align*}
	
	By combining the previous results we obtain
	{\footnotesize
		\begin{align*}
		&-\log q(f) = \frac{1}{2} \log \abs[\Sigma] + \frac{1}{2} (f-\mu)^T \Sigma^{-1} (f-\mu) \\
		&= \frac{1}{2} \log \frac{\abs[S]}{\abs[K_{uu}]} + \frac{1}{2} (f-\mu)^T (I - \Psi_u ( K_{uu}^{-1} - S^{-1} ) \Psi_u^T) (f-\mu) \\
		&= \frac{1}{2} \log \frac{\abs[S]}{\abs[K_{uu}]} + \frac{1}{2} (f-\mu)^T ( N_u + \Psi_u S^{-1} \Psi_u^T) (f-\mu) \\
		&= \frac{1}{2} \log \frac{\abs[S]}{\abs[K_{uu}]} + \frac{1}{2} f_\perp^T N_u f_\perp + \frac{1}{2}(f_\parallel-\mu)^T (  \Psi_u S^{-1} \Psi_u^T) (f_\parallel-\mu) \\
		&= \frac{1}{2} \log \frac{\abs[S]}{\abs[K_{uu}]} + \frac{1}{2} f_\perp^T N_u f_\perp \\
		&+ \frac{1}{2}(f_u-m)^T K_{uu}^{-1}\Psi_u^T \Psi_u S^{-1} \Psi_u^T\Psi_u K_{uu}^{-1} (f_u-m) \\
		&= \frac{1}{2} \log \frac{\abs[S]}{\abs[K_{uu}]} + \frac{1}{2} b^T \Psi_X^T N_u \Psi_X b + \frac{1}{2}(f_u-m)^T S^{-1} (f_u-m).
		\end{align*}
	}
	We further note that $N_u = I - \Psi_u K_{uu}^{-1}\Psi_u^T$ and $\Psi_X^T (I - \Psi_u K_{uu}^{-1}\Psi_u^T) \Psi_X = K_X - \hat{K_u}$, therefore we have
	{\footnotesize
		\begin{align*}
		-\log q(f) &= \frac{1}{2} \log \frac{\abs[S]}{\abs[K_{uu}]} + \frac{1}{2} b^T ( K_X - \hat{K_u}) b \\
		&\quad+ \frac{1}{2}(f_u-m)^T S^{-1} (f_u-m) \\
		&= \frac{1}{2} \log \frac{\abs[S]}{\abs[K_{uu}]} + \frac{1}{2} b^T \hat{N}^T ( K_X - \hat{K_u})\hat{N} b \\
		&\quad+ \frac{1}{2}(f_u-m)^T S^{-1} (f_u-m) \\
		&= \frac{1}{2} \log \frac{\abs[S]}{\abs[K_{uu}]} + \frac{1}{2}(f_u-m)^T S^{-1} (f_u-m) \\
		&\quad+ \frac{1}{2} b^T \hat{N}^T ( K_X - \hat{K_u})( K_X - \hat{K_u})^{-1}( K_X - \hat{K_u})\hat{N} b \\
		&= \frac{1}{2} \log \frac{\abs[S]}{\abs[K_{uu}]} + \frac{1}{2}(f_u-m)^T S^{-1} (f_u-m) \\
		&\quad+ \frac{1}{2}  ( f_X - \Psi_X K_{uu}^{-1} f_u)^T( K_X - \hat{K_u})^{-1}( f_X - \Psi_X K_{uu}^{-1} f_u) \\
		&= \frac{1}{2} (\log \abs[S] -\log\abs[K_{uu}] + \log\abs[K_X - \hat{K_u}] - \log\abs[K_X - \hat{K_u}] ) \\
		&\quad+\frac{1}{2}  ( f_X - \Psi_X K_{uu}^{-1} f_u)^T( K_X - \hat{K_u})^{-1}( f_X - \Psi_X K_{uu}^{-1} f_u) \\
		&\quad+ \frac{1}{2}(f_u-m)^T S^{-1} (f_u-m) \\
		&= \frac{1}{2} ( -\log\abs[K_{uu}] - \log\abs[K_X - \hat{K_u}] ) \\
		&\quad- \log p(f_X \mid f_u) -\log q(f_u)
		\end{align*}
	}
\end{proof}

\begin{proof}[Proof of equation~\eqref{eq:lowBoundDecomp}]
	Recall that 
	{\small
		\begin{align*}
		\max p(y) \geq \max &\int q(f_X \mid f_u) \log \frac{p(y\mid f_X) p(f_X \mid f_u) p(f_u)}{q(f_X, f_u)} \\
		&= \mathcal{L}(q(f_X, f_u)).
		\end{align*}
	}
	Now by using Remark~\ref{rem:equivCoupled} we can write $\mathcal{L}$ as a function of $q(f)$. 
	\begin{align*}
	\mathcal{L}(q(f)) &= \int q(f) \log \frac{p(y \mid f) p(f)}{q(f)} df \\
	&= \int p(f_X \mid f_u) q(f_u) \abs[K_{uu}]^{1/2} \abs[K_X - \hat{K_u}]^{1/2} \\
	&\cdot\log \frac{p(y \mid f) p(f_X \mid f_u) p(f_u) \abs[K_{uu}]^{1/2} \abs[K_X - \hat{K_u}]^{1/2}}{p(f_X \mid f_u) q(f_u) \abs[K_{uu}]^{1/2} \abs[K_X - \hat{K_u}]^{1/2}} df \\
	&= \int p(f_X \mid f_u) q(f_u) \log \frac{p(y \mid f_X) p(f_X \mid f_u) p(f_u)}{p(f_X \mid f_u) q(f_u)} \\
	&\cdot\abs[K_{uu}]^{1/2} df_\parallel \abs[K_X - \hat{K_u}]^{1/2} df_\perp \\
	&=\int p(f_X \mid f_u) q(f_u) \log \frac{p(y \mid f) p(f_X \mid f_u) p(f_u)}{p(f_X \mid f_u) q(f_u)} df_X df_u \\ 
	&= \int q(f_X , f_u) \log \frac{p(y \mid f_X) p(f_X, f_u)}{q(f_X, f_u)} df_X df_u \\
	&= \mathcal{L}(q(f_X,f_u))
	\end{align*}	
\end{proof}

\section{Gaussian likelihood case}
\label{sec:Gausslike}

In the regression case with Gaussian likelihood, i.e.~$p(y \mid f) \sim N(y \mid f, \sigma_n^2 I_N)$, we can derive analytical expressions for the variational mean and covariance. Recall that if $q(f) \sim N(f \mid \mu, \Sigma)$, with $\mu, \Sigma$ defined as in~\eqref{eq:orthDecVFFmean} and~\eqref{eq:orthDecVFFvar} respectively, then the predictive distribution $q(f(\xb))$ is normal with mean and variances given by
\begin{align*}
m(\xb) &= (k_{\xb,\gamma} -k_{\xb,\beta} K_\beta^{-1} k_{\beta, \gamma} )a_\gamma +k_{\xb,\beta} a_\beta = K_{\xb,\alpha} a_\alpha \\
s(\xb) &= k(\xb,\xb) - k_{\xb,\beta}K_\beta k_{\beta,\xb}  + k_{\xb,\beta}K_\beta^{-1} S K_\beta^{-1} k_{\beta,\xb},
\end{align*}
where $K_{\xb,\alpha} = [K_{\xb,\gamma},K_{\xb,\beta}]$ and $a_\alpha = \begin{bmatrix}
a_\gamma \\
a_\beta - K_\beta^{-1} K_{\beta,\gamma} a_\gamma
\end{bmatrix}$.

We can plug-in the predictive distribution in the evidence lower bound 
\begin{equation*}
\mathcal{L}(q) = \E_{q(f(\xb))}[\log p\left(y \mid  f(\xb) \right)] - KL[ q \parallel p ],
\end{equation*}
which is analytical since the likelihood is Gaussian. By computing the derivatives with respect to $a_\alpha$ and $S$ and by setting them to zero we obtain 
\begin{align*}
S &= (K_\beta^{-1}+ \tfrac{1}{\sigma_n^2} K_\beta^{-1} K_{\beta,\xb} K_{\xb, \beta}K_\beta^{-1})^{-1}  \\
&= K_\beta(K_\beta+ \tfrac{1}{\sigma_n^2} K_{\beta,\xb} K_{\xb, \beta})^{-1}K_\beta \\
a_\alpha &= ( K_{\alpha, \xb} K_{\xb, \alpha} +  \sigma_{n}^2K_{\alpha})^{-1} K_{\alpha, \xb} \yb
\end{align*}
where $K_\alpha = \begin{bmatrix}
K_\gamma & K_{\gamma,\beta} \\
K_{\beta,\gamma} & K_{\beta}
\end{bmatrix}$. 

See section~\ref{subsec:ELBOanalytical} for the detailed calculations.

\subsection{Expected log-likelihood}

Consider $q(f) \sim N(f \mid \mu, \Sigma)$ with $\mu$ and $\Sigma$ as in~\eqref{eq:orthDecVFFmean} and~\eqref{eq:orthDecVFFvar} respectively, then for $x \in \inSpace$ the predictive distribution is $q(f(x))\sim N(f(x) \mid m(x), s(x) )$
\begin{align*}
m(x) &= (k_{x,\gamma} -k_{x,\beta} K_\beta^{-1} k_{\beta, \gamma} )a_\gamma +k_{x,\beta} a_\beta \\
s(x) &= k(x,x) - k_{x,\beta}K_\beta k_{\beta,x}  + k_{x,\beta}K_\beta^{-1} S K_\beta^{-1} k_{\beta,x} 
\end{align*}

\subsection{ELBO}
\label{subsec:ELBOanalytical}

Recall that the ELBO is
\begin{equation}
\mathcal{L}(q) = \E_{q(f(x))}\left[\log p\left(y \mid f(x)\right)\right] - KL[q(f) \parallel p(f)] 
\label{eq:elbo1}
\end{equation}
where $q(f(x)) \sim N(f(x) \mid m(x), s(x))$, $p(y \mid f(x)) \sim N(y \mid f(x), \sigma^2_{n})$. We can then develop equation~\eqref{eq:elbo1} as follows.
{\small
	\begin{align*}
	\mathcal{L}(q) &= \sum_{i=1}^N \int \log p\left(y_i \mid f(x_i)\right) N(f(x_i) \mid m(x_i), s(x_i)) df(x_i) \\
	&- KL[q(f) \parallel p(f)] \\
	&= \int \big(-N\log(2\pi \sigma_{n}^2) - \tfrac{1}{2\sigma_{n}^2}\yb^T\yb + \tfrac{2}{2\sigma_{n}^2}\yb^T \fb \\
	&- \tfrac{1}{2\sigma_{n}^2}\fb^T \fb\big) N(\fb \mid m(x), s(x)) d\fb - KL[q(f) \parallel p(f)] \\
	&= -N\log(2\pi \sigma_{n}^2) - \tfrac{1}{2\sigma_{n}^2}\yb^T\yb + \tfrac{2}{2\sigma_{n}^2}\yb^T m(\xb) \\
	&- \tfrac{1}{2\sigma_{n}^2}m(\xb)^T m(\xb) -\tfrac{1}{2\sigma_{n}^2}s(\xb)  - KL[q(f) \parallel p(f)] \\
	&= -N\log(2\pi \sigma_{n}^2) - \tfrac{1}{2\sigma_{n}^2}\yb^T\yb + \tfrac{2}{2\sigma_{n}^2}\yb^T K_{\xb, \alpha} a_\alpha -\tfrac{1}{2\sigma_{n}^2}K_{\xb,\xb} \\
	&- \tfrac{1}{2\sigma_{n}^2}K_{\xb,\beta} K_\beta^{-1} (S-K_\beta) K_\beta^{-1} K_{\beta,\xb} \\
	&- \tfrac{1}{2\sigma_{n}^2} a_\alpha^T K_{\alpha, \xb} K_{\xb, \alpha} a_\alpha  \\
	&- \frac{1}{2}\bigg( tr(S K_\beta^{-1} ) -M_\beta +a_\alpha^T K_\alpha a_\alpha -\ln \left( \abs[ K_\beta^{-1}S] \right) \bigg).
	\end{align*}
}
where $K_{\xb,\alpha} = [K_{\xb, \gamma} \ K_{\xb,\beta}]$, $a_\alpha = \begin{bmatrix}
a_\gamma \\
a_\beta - K_\beta^{-1} K_{\beta, \gamma} a_\gamma
\end{bmatrix}$ and $K_\alpha = \begin{bmatrix}
K_\gamma & K_{\gamma, \beta} \\
K_{\beta,\gamma} & K_{\beta} \\
\end{bmatrix}$. 
By taking the derivatives with respect to $S$ and $a_\alpha$ we obtain
\begin{align*}
\frac{\partial \mathcal{L}}{\partial S} = -\tfrac{1}{2} K_\beta^{-1} +\tfrac{1}{2}S^{-1}- \tfrac{1}{2\sigma_n^2} K_\beta^{-1} K_{\beta,\xb} K_{\xb, \beta}K_\beta^{-1} = 0 \\
\frac{\partial \mathcal{L}}{\partial a_\alpha} = \tfrac{1}{\sigma_n^2} K_{\alpha, \xb} \yb - \tfrac{2}{2\sigma_{n}^2} K_{\alpha, \xb} K_{\xb, \alpha} a_\alpha -  K_{\alpha} a_\alpha =0
\end{align*}
which results in 
\begin{align*}
S &= (K_\beta^{-1}+ \tfrac{1}{\sigma_n^2} K_\beta^{-1} K_{\beta,\xb} K_{\xb, \beta}K_\beta^{-1})^{-1}  \\
&= K_\beta(K_\beta+ \tfrac{1}{\sigma_n^2} K_{\beta,\xb} K_{\xb, \beta})^{-1}K_\beta \\
a_\alpha &= ( K_{\alpha, \xb} K_{\xb, \alpha} +  \sigma_{n}^2K_{\alpha})^{-1} K_{\alpha, \xb} \yb
\end{align*}

\section{Non-stationary kernels}

The ODVFF method is built on RKHS features which are only defined for a few stationary kernels, as explained in Section~\ref{subsec:interDomain}. Nonetheless it is possible to adapt the method for kernels which are sums of a stationary kernel plus a non-stationary one. Consider a stationary kernel $k^{(S)}$ and a non-stationary kernel $k^{(NS)}$. We assume that the GP kernel be decomposed in an additive combination of a stationary and a non-stationary kernel. We can write it as $f(\xb) = f^{(S)}(\xb)+f^{(NS)}(\xb)$, $\xb \in \inSpace$, where $f^{(S)} \sim GP(0, k^{(S)}(\xb,\xb^\prime))$ and $f^{(NS)} \sim GP(0, k^{(NS)}(\xb,\xb^\prime))$. This results in an overall process defined as
\begin{equation*}
f \sim GP\left(0,k^{(S)}(\xb,\xb^\prime) + k^{(NS)}(\xb,\xb^\prime)\right).
\end{equation*}
We can then split the covariance parametrization in a stationary and in a non-stationary part. For the stationary part we can select a Mat\'ern kernel and define $\abs[\beta]/2$ features
\begin{equation*}
\zeta_{i} = <\phi_i, f^{(S)}>_{\mathcal{H}} \quad i=1, \ldots, \abs[\beta]/2,
\end{equation*}
where $\mathcal{H}$ is the RKHS associated with the stationary kernel. For the non-stationary part we can fall back on the inducing point framework and select $\abs[\beta]/2$ inducing points $\gamma_i$, $i=1, \ldots, \abs[\beta]/2$.   The Gram matrix $K_{\phi,\phi}$ is then the $\abs[\beta] \times \abs[\beta]$ block-diagonal matrix where the first block, of dimension $\abs[\beta]/2 \times \abs[\beta]/2$, is the Gram matrix associated with the features $\zeta_i$ while the second block (also of dimension $\abs[\beta]/2 \times \abs[\beta]/2$) is the Gram matrix associated with the inducing point basis. Analogously we can adapt the vector $\Psi_{\beta,\gamma}$ as 
\begin{align*}
\Psi_{\beta\gamma} &= [\operatorname{Cov}(\beta_1,f^{(S)}(\cdot)), \ldots, \operatorname{Cov}(\beta_{\abs[\beta]/2},f^{(S)}(\cdot)), \\
&\qquad k^{(NS)}(r_{\gamma_1},\cdot),\ldots, k^{(NS)}(r_{\gamma_{\abs[\beta]/2}},\cdot)]^T  \\
&= [ \phi^{(S)}_1(\cdot), \ldots, \phi^{(S)}_{\abs[\beta]/2}(\cdot),\\
&\qquad k^{(NS)}(r_{\gamma_1},\cdot),\ldots, k^{(NS)}(r_{\gamma_{\abs[\beta]/2}},\cdot)]^T,
\end{align*}
where $r_{\gamma_1}, \ldots, r_{\gamma_{\abs[\beta]/2}}$ are the positions of the inducing points $\gamma_1, \ldots, \gamma_{\abs[\beta]/2}$.

\begin{figure}
	\includegraphics[width=\linewidth]{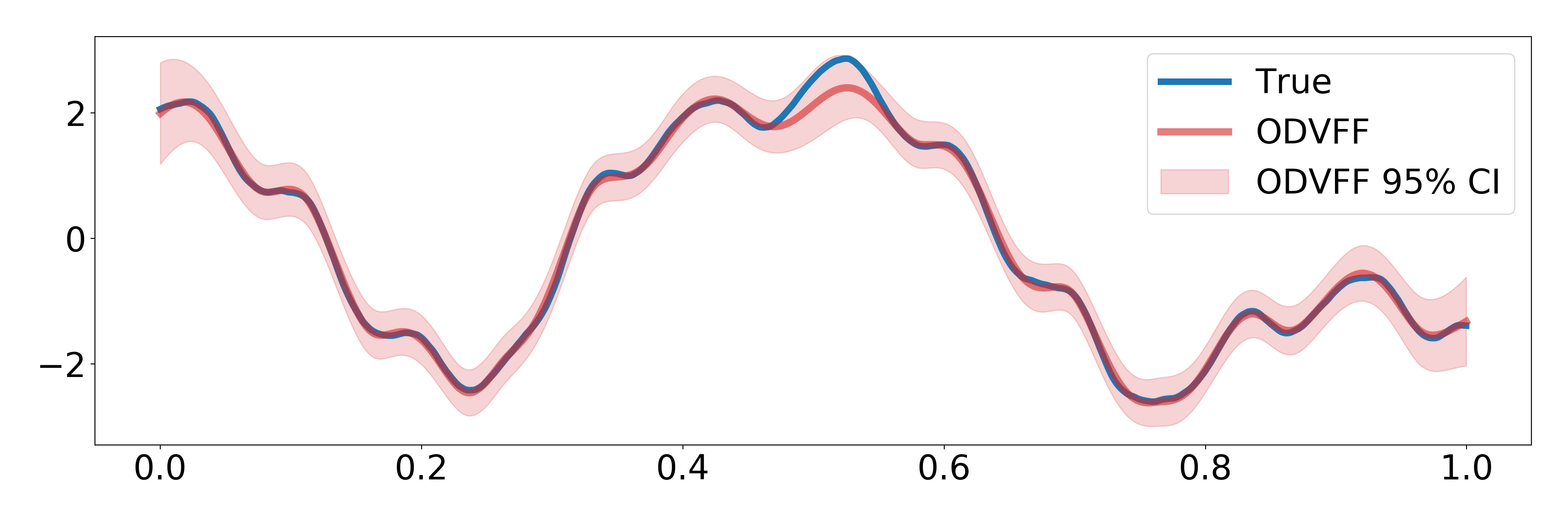}
	\caption{ODVFF with additive non-stationary kernel.}
	\label{fig:1dmultiKernel}
\end{figure}

Figure~\ref{fig:1dmultiKernel} shows an example where data generated from a GP with an additive kernel sum of a Matern ($\nu=3/2$) and periodic kernel. We used a period of $0.5$ and lengthscales of $0.2$ and generated $1000$ training data points from a realization of the GP. We tested on $3000$ data points generated from the same realization. The ODVFF obtains a test log-likelihood of $0.45$ and a test RMSE of $0.14$. For reference a full GP implementation obtains a test log-likelihood of $0.78$ and a tRMSE of $0.11$. 

\section{Further experimental results}

\subsection{Synthetic data}
\label{subsec:addSynthData}

In this section we report the results for the experiments on synthetic data introduced in Section~\ref{sec:experiments}, main text. 

\begin{table}
	\begin{minipage}{0.48\textwidth}
	\caption{Synthetic data sets with $D=1$. Test RMSE values, lower values denote a better fit.}
	\label{tab:synthRMSE_1d}
	\begin{center}
		\footnotesize
		\begin{tabular}{lrrrr}
			\toprule
			& \multicolumn{2}{c}{$N=50000$} & \multicolumn{2}{c}{$N=100000$} \\
			\cmidrule(lr){2-3}
			\cmidrule(lr){4-5}
			&  $\abs[\beta]=10$ &        $\abs[\beta]=50$ &        $\abs[\beta]=10$ &        $\abs[\beta]=50$ \\
			
			\midrule
			ODVFF &  \textbf{0.221} &  \textbf{0.214} &  0.228 &  \textbf{0.210} \\
			ODVGP &  0.224 &  \textbf{0.214} &  \textbf{0.221} &  0.212 \\
			SVGP  &  0.242 &  0.242 &  0.226 &  0.226 \\
			SGPR  & \multicolumn{2}{c}{0.212} &    \multicolumn{2}{c}{$-$}  \\
			\bottomrule
		\end{tabular}
	\end{center}
\end{minipage} \hfill
\begin{minipage}{0.48\textwidth}
	\caption{Synthetic data sets with $D=1$. Test mean coverage values, theoretical value $95\%$.}
	\label{tab:synthMcov_1d}
	\begin{center}
		\footnotesize
		\begin{tabular}{lrrrr}
			\toprule
			& \multicolumn{2}{c}{$N=50000$} & \multicolumn{2}{c}{$N=100000$} \\
			\cmidrule(lr){2-3}
			\cmidrule(lr){4-5}
			&        $\abs[\beta]=10$ &        $\abs[\beta]=50$ &        $\abs[\beta]=10$ &        $\abs[\beta]=50$ \\
			
			\midrule
			ODVFF &  0.954 &  0.960 &  0.950 &  0.870 \\
			ODVGP &  0.945 &  0.944 &  0.946 &  0.830 \\
			SVGP  &  0.939 &  0.939 &  0.944 &  0.944 \\
			SGPR  &  \multicolumn{2}{c}{0.949} &    \multicolumn{2}{c}{$-$} \\
			\bottomrule
		\end{tabular}
	\end{center}
\end{minipage}
\end{table}

\begin{table}
	\begin{minipage}{0.48\textwidth}
	\caption{Synthetic data sets with $D=5$. Test RMSE values, lower values denote a better fit.}
	\label{tab:synthRMSE_5d}
	\begin{center}
		\footnotesize
		\begin{tabular}{lrrrr}
			\toprule
			& \multicolumn{2}{c}{$N=50000$} & \multicolumn{2}{c}{$N=100000$} \\
			\cmidrule(lr){2-3}
			\cmidrule(lr){4-5}
			&        $\abs[\beta]=10$ &        $\abs[\beta]=50$ &        $\abs[\beta]=10$ &        $\abs[\beta]=50$ \\
			
			\midrule
			ODVFF &  1.320 &  1.324 &  1.306 &  1.309 \\
			ODVGP &  1.316 &  1.317 &  \textbf{1.304} &  1.305 \\
			SVGP  &  \textbf{1.316} &  \textbf{1.316} &  \textbf{1.304} &  \textbf{1.304} \\
			SGPR  &  \multicolumn{2}{c}{1.316} &   \multicolumn{2}{c}{$-$} \\
			\bottomrule
		\end{tabular}
	\end{center}
\end{minipage} \hfill
\begin{minipage}{0.48\textwidth}
	\caption{Synthetic data sets with $D=5$. Test mean coverage values, theoretical value $95\%$.}
	\label{tab:synthMcov_5d}
	\begin{center}
		\footnotesize
		\begin{tabular}{lrrrr}
			\toprule
			& \multicolumn{2}{c}{$N=50000$} & \multicolumn{2}{c}{$N=100000$} \\
			\cmidrule(lr){2-3}
			\cmidrule(lr){4-5}
			& $\abs[\beta]=10$ &        $\abs[\beta]=50$ &        $\abs[\beta]=10$ &        $\abs[\beta]=50$ \\
			
			\midrule
			ODVFF &  0.700 &  0.544 &  0.471 &  0.542 \\
			ODVGP &  0.685 &  0.532 &  0.412 &  0.446 \\
			SVGP  &  0.631 &  0.631 &  0.442 &  0.442 \\
			SGPR  &  \multicolumn{2}{c}{0.953} & \multicolumn{2}{c}{$-$} \\
			\bottomrule
		\end{tabular}
	\end{center}
\end{minipage}
\end{table}

\begin{table}
	\begin{minipage}{0.48\textwidth}
	\caption{Synthetic data sets with $D=10$. Test RMSE values, lower values denote a better fit.}
	\label{tab:synthRMSE_10d}
	\begin{center}
		\footnotesize
		\begin{tabular}{lrrrrrr}
			\toprule
			& \multicolumn{2}{c}{$N=50000$} & \multicolumn{2}{c}{$N=100000$} \\
			\cmidrule(lr){2-3}
			\cmidrule(lr){4-5}
			& $\abs[\beta]=10$ &        $\abs[\beta]=50$ &        $\abs[\beta]=10$ &        $\abs[\beta]=50$ \\
			
			\midrule
			ODVFF &  1.536 &  1.537 &  1.556 &  1.554 \\
			ODVGP &  1.526 &  1.527 &  1.547 &  1.548 \\
			SVGP  &  \textbf{1.526} &  \textbf{1.526} &  \textbf{1.547} &  \textbf{1.547} \\
			SGPR  & \multicolumn{2}{c}{1.526} &  \multicolumn{2}{c}{$-$}  \\
			\bottomrule
		\end{tabular}
	\end{center}
\end{minipage} \hfill
\begin{minipage}{0.48\textwidth}
	\caption{Synthetic data sets with $D=10$. Test mean coverage values, theoretical value $95\%$.}
	\label{tab:synthMcov_10d}
	\begin{center}
		\footnotesize
		\begin{tabular}{lrrrr}
			\toprule
			& \multicolumn{2}{c}{$N=50000$} & \multicolumn{2}{c}{$N=100000$} \\
			\cmidrule(lr){2-3}
			\cmidrule(lr){4-5}
			& $\abs[\beta]=10$ &        $\abs[\beta]=50$ &        $\abs[\beta]=10$ &        $\abs[\beta]=50$ \\
			
			\midrule
			ODVFF &  0.653 &  0.719 &  0.708 &  0.766 \\
			ODVGP &  0.629 &  0.422 &  0.618 &  0.364 \\
			SVGP  &  0.555 &  0.555 &  0.543 &  0.543 \\
			SGPR  &  \multicolumn{2}{c}{0.948}  &   \multicolumn{2}{c}{$-$}  \\
			\bottomrule
		\end{tabular}
	\end{center}
\end{minipage}
\end{table}

\subsection{Classification example}

The orthogonally decoupled variational Fourier feature method was applied here only in regression tasks, however it is also possible to apply the method to classification tasks. In fact in the formulation described in Sections~\ref{sec:ODVFF} and \ref{sec:sparseRKHS}, main text, we can use any likelihood function $p(\yb \mid \fb)$. In this example we consider input data $\Xb \in \R^{N \times D}$ and an output vector $\yb = \{y_i\}_{i = 1, \ldots, N}$ where $y_i = \{ -1,1\}$. We assume that the labels are assign as $y_i = \operatorname{sign}(f(x_i) +\epsilon_i)$, where $f \sim GP(0,k)$ and $\epsilon_i$ are independent standard Gaussian noise. We consider a probit likelihood $p(\yb \mid \fb) = \prod_{i=1}^{N}F(y_i f_i)$ where $F(\cdot)$ is the c.d.f. of a standard Gaussian random variable. We can train our ODVFF method by learning the variational distribution $q$ as in the regression case. 

\begin{figure*}
	\centering
	\includegraphics[width=\textwidth]{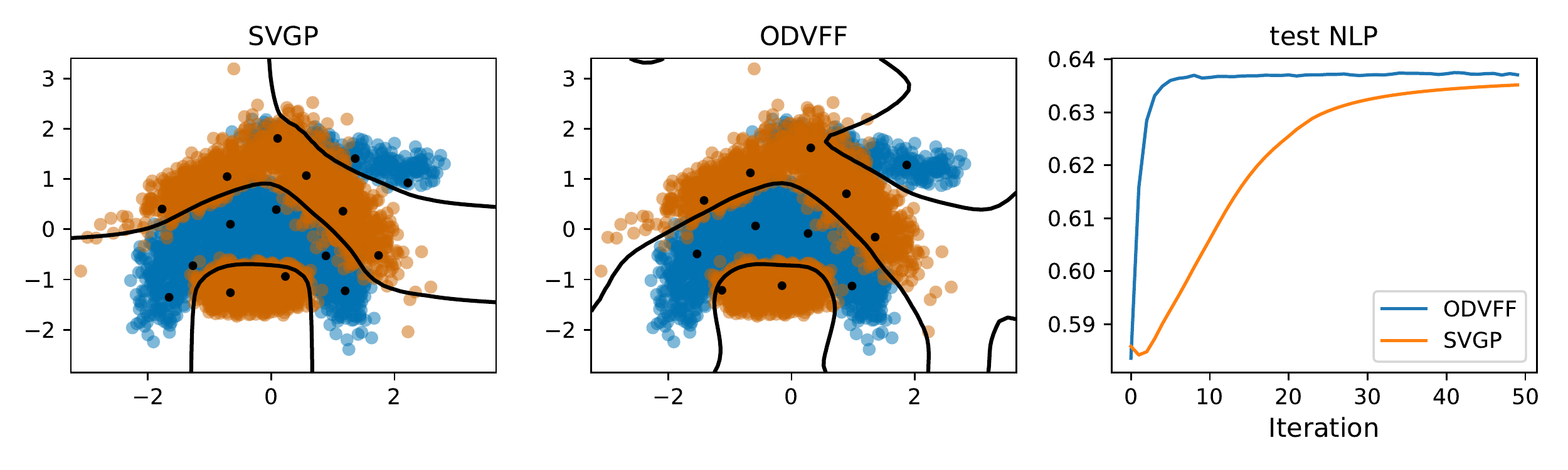}
	\caption{Comparison of SVGP and ODVFF on Banana dataset.}
	\label{fig:banana1}
\end{figure*}

As an example we consider here the banana dataset in \citet{Hensman.eal15} and we train ODVFF and SVGP with $\abs[\alpha]=12$ and $\abs[\beta]=4$. We train the models with $5000$ iterations and mini-batch size $100$. Figure~\ref{fig:banana1} shows the decision boundary obtained with SVGP and with ODVFF. We notice how both decision boundaries closely resemble the full GP decision boundary. Here the ODVFF method was run with Kronecker product covariance kernel, however as $\abs[\beta]$ and $D$ increase this method is not usable in practice. Here the alternative of using additive covariances results in very low accuracy, so it is also not practically viable. Alternative implementations such as the product of two  Kronecker covariances described in \citet{Hensman.etal2018} was not explored here. Nonetheless additional work is needed to make such methods more scalable in the input dimension.

\end{document}